\documentclass[10pt, conference, compsocconf]{IEEEtran}
\usepackage{cite}
\usepackage{float}
\ifCLASSINFOpdf

\else

\fi

	\usepackage{titlesec}
	\usepackage{makecell}
	\usepackage{adjustbox}
	\usepackage{amsmath} 
	\usepackage{amssymb}
	\usepackage{tabularx}
	\usepackage{tikz}
	\usepackage{placeins}
	\usepackage{booktabs}
	\usepackage[justification=raggedright]{caption}
	\usetikzlibrary{shapes.geometric, arrows, chains, decorations.markings, positioning, fit, calc}
	
	\titlespacing*{\subsubsection}{0pt}{*0.2}{*0.2}
	\usepackage{booktabs}
	\usepackage{pifont}  
	
	\newcommand{\cmark}{\ding{51}} 
	\newcommand{\xmark}{\ding{55}} 

\begin{document}
		%
		\title{Human Action Recognition from Point Clouds over Time}

		
		\author{\IEEEauthorblockN{}
			\IEEEauthorblockA{
				James Dickens\\
				University of Ottawa\\
				Ontario, Canada\\
				jdick088@uottawa.ca
		}}

		\maketitle

		\begin{abstract}
			Recent research into human action recognition (HAR) has focused predominantly on skeletal action recognition and video-based methods. With the increasing availability of consumer-grade depth sensors and Lidar instruments, there is a growing opportunity to leverage dense 3D data for action recognition, to develop a third way. This paper presents a novel approach for recognizing actions from 3D videos by introducing a pipeline that segments human point clouds from the background of a scene, tracks individuals over time, and performs body part segmentation. The method supports point clouds  from both depth sensors and monocular depth estimation. At the core of the proposed HAR framework is a novel backbone for 3D action recognition, which combines point-based techniques with sparse convolutional networks applied to voxel-mapped point cloud sequences. Experiments incorporate auxiliary point features including surface normals, color, infrared intensity, and body part parsing labels, to enhance recognition accuracy. Evaluation on the NTU RGB-D 120 dataset demonstrates that the method is competitive with existing skeletal action recognition algorithms. Moreover, combining both sensor-based and estimated depth inputs in an ensemble setup, this approach achieves 89.3\% accuracy when different human subjects are considered for training and testing, outperforming previous point cloud action recognition methods.

		\end{abstract}


		\IEEEpeerreviewmaketitle

		\section{Introduction}
		Within the canon of computer vision literature, human action recognition has been the focus of a wide breadth of research. Its applications are diverse, with algorithms designed to recognize human motion and behavior forming the foundation for many consumer products. In surveillance, action recognition enables automated anomaly and violence detection, in addition to identifying falls in elderly persons \cite{FallDetection}. It also facilitates automated video annotation, allowing videos to be analyzed without requiring human intervention—a capability particularly valuable in sports analysis \cite{AutomatedTimeStamping}. Moreover, in autonomous driving, understanding human actions is crucial to ensure the safety of pedestrians and drivers alike \cite{AutonomousDriving}.\\
		\indent Recently, deep-learning based approaches have largely dominated research into human action recognition, leveraging large-scale video datasets to train discriminative models capable of distinguishing an ever-increasing set of human actions. Two popular streams of models have been developed, based on raw video inputs and keypoints over time, with the latter commonly referred to as skeletal action recognition. Video-based action recognition has been successfully demonstrated with convolutional neural networks \cite{I3D, SlowFast, X3D} and vision transformers \cite{MVit, SWinVid}, but suffers from the need to reduce frame sizes with interpolation due to memory concerns, with a resolution size per frame of $224 \times 224$ being common, limiting their use for fine-grained actions or multi-person scenarios in high resolution videos. Further, video-based models require extensive compute requirements in terms of training time and inference.\\
		\indent The skeletal action recognition approach uses sequences of keypoints detected on the human body as input, with 3D coordinates obtained by projection using depth images or estimation techniques. In this domain, graph convolutional networks \cite{ST-GCN, 2SAGCN, DGCN} have shown great success at recognizing actions from sequences of keypoints, however they have limitations. Keypoints require an estimation stage that is subject to error from jitter, occlusion from objects and other people in the scene, self-occlusion, and projection errors in the case of noisy depth values \cite{PoseFix}. Moreover, popular keypoint estimators such as Yolov8, AlphaPose and OpenPose \cite{Yolov8, AlphaPose, OpenPose} output unique joint topologies, limiting the use of transfer learning across keypoint data extracted with different algorithms. Additionally, many keypoint estimators offer limited or inaccurate representations of hand keypoints which are crucial for recognizing many actions.\\
		\indent With the proliferation of consumer grade depth sensors such as Microsoft's Kinect series (v1, v2, Azure), as well as Lidar-endowed robots and instruments, researchers have begun to develop methods for human action recognition based on three dimensional video data, sampling a fixed number of points per frame as input, discussed in Section II-B. These approaches largely draw from the existing modern 3D computer vision literature concerning point cloud classification, segmentation, and object detection. Existing methods do not segment the individuals from the background beyond simple depth thresholding or histogram techniques, and rely purely on point-based approaches that require extensive local spatio-temporal neighborhood grouping computations to group points in both space and time. Further, they require a depth sensor or multi-view camera setups to obtain depth information, which is not practical or available in many settings. Recent research into monocular depth estimation \cite{ZoeDepth, DepthAnything}, providing the ability to estimate metric or ordinal disparity maps from images and videos, now offers the promise of estimated point clouds, which are explored in this work. \\
		\indent In light of these limitations, in this work, a novel pipeline is developed for two scenarios. 
		\begin{itemize}
			\item Depth sensing technology is present, in which case real world point clouds sequences are considered. 
			\item An RGB video is used as input, in which case modern monocular depth estimation algorithms are used to obtain estimated point clouds.
		\end{itemize}
		In each scenario, the actors in a sequence are segmented from their background, assigned tracking identification, and a human body part segmentation algorithm is applied, with light instance denoising applied both in image space and 3D.\\
		\indent Inspired by recent research into 3D object detection based on voxelized point clouds, alongside point-based approaches, a novel backbone for human action recognition is developed. The proposed approach integrates sparse convolution and point-wise embeddings. The input to this model consists of segmented point clouds, which are obtained through the proposed pipeline.   
		
		\section{Related Work}
		\subsection{Point Cloud Deep Learning}
		As noted by Lu et al. \cite{PointCloudSurveys}, the focus of modern point cloud deep learning has largely been model classification, part segmentation, semantic segmentation, and 3D object detection and tracking. Other topics of interest include point cloud registration, learnable downsampling and upsampling, denoising, and scene completion from occluded inputs. Point cloud methods can be distinguished by two main tracks. The first track are those approaches that employ purely \textit{point-based} operations to learn features from localized aggregation operators. Neighborhoods are estimated using K-nearest neighbors (K-nn), ball radius queries, or windows of points serialized with space-filling curves. A second track maps points to 3D grid locations, known as voxels, and uses sparse convolution or attention operations based on non-empty regions to compute features. In this section, point cloud models that employ projection to images are not considered, but rather representative works in both tracks are summarized. \vspace{5px}
	 \subsubsection{Point-Based Models} 
	 \vspace{5px} The seminal work of PointNet \cite{PointNet} introduced by Qi et al. developed a set-based approach to point cloud classification and part-segmentation. Points are mapped to a cannonical embedding space with multiple layers of learnable T-Net (Transformation Net) operations, using max pooling as a set aggregation tool, with individual point-wise features learned by multi-layer perceptrons (MLPs). The follow-up work PointNet++ \cite{PointNet++} introduced the novel use of neighborhood grouping by radius search for local feature aggregation, in addition to the use of multi-scale (in terms of points) and multi-radius feature learning, where points are successively downsampled along the network according to the iterative farthest point sampling algorithm (IFPS). Dynamic Graph CNN \cite{DGCNN} constructs dynamic K-nn graphs based on spatial proximity alongside point-wise features of neighborhood points for local feature aggregation. \\
  \indent In keeping with the trend in modern computer vision to explore the use of concepts from the Transformer architecture \cite{Transformer}, PointTransformer v1 \cite{PointTransformerv1} makes use of self-attention mechanisms to aggregate features in local radii around points, using relative positional embeddings, employed for both point cloud classification and segmentation. The follow-up model PointTransformer v2 \cite{PointTransformerv2} utilizes channel groupings in the attention mechanism, and uniform grid-based pooling in the down-sampling and upsampling stages of the network. The most recent PointTransformer v3 \cite{PointTransformerv3} focuses on large-scale semantic segmentation by using point serialization with space-filling curves (z-order curves and Hilbert curves), to learn features within windows of 1 dimensional arrays of points, introducing serialized pooling/unpooling. They leverage serialization to avoid memory-expensive neighborhood searches. In a similar vein, Wang develops the OctFormer model \cite{OctFormer}, which also incorporates z-order curves to reshape points into windows, computing attention within windows, but computes conditional positional encoding of points using octree-based convolution. 
	 \subsubsection{Voxel-Based Models} 
	 Voxel-based models are popular within the literature concerning 3D object detection and tracking from point clouds, where points are assigned to a 3D grid location. Within these grid cells, features are aggregated using the mean/max of their channel values, often using normalized 3D locations and other features such as intensity, color, and surface normals. One of the first works in this category, VoxNet \cite{VoxNet} used a 3D convolutional neural network input with 3D occupancy grids, using ray tracing to compute the number of hits per voxel given a camera view. Using a sliding box or pre-segmented input, 3D object detection can be performed in real-time with this approach. \\
	 \indent Data in 3D is often full of large connected empty regions, consequently a set of architectures known as sparse convolutional neural networks, or sparse CNNs, has been developed to increase the efficiency of 3D convolution. Yan et al. introduce the SECOND (Sparsely Embedded Convolutional Detection) model for 3D object detection \cite{SECOND}, and define sparse convolution in 3D, outlining a strategy for parallelization with CUDA kernels executed on the GPU. In this formulation of convolution, only voxels whose local neighborhood has non-empty points defined by the extent of a kernel are convolved.\\ \indent  Choy et al. develop Minkowski Networks \cite{MinkowskiEngine}, and the corresponding software library the Minkowski Engine, in which sparse convolution is extended to non-tesseract kernels, adding an additional variant known as sub-manifold sparse convolution \cite{SubManifoldConvolution}. In sub-manifold sparse convolution, only input voxels, referred to as sites, whose kernel centers are non-empty are convolved, fully preserving sparsity. This contrasts the de-sparsification effect that occurs with regular sparse convolution, in which empty sites become non-empty if their neighborhoods are non-empty. Minkowski networks are applied to the task of 4D semantic segmentation, and voxel-based 3D model classification. Chen et al. research focal sparse convolution \cite{FocalSparseConv}, a specialized sparse convolutional layer to predict importance weights related to de-sparsified sites.  Thresholding learned by proximity to real object bounding boxes is employed in order to prune sites for improved efficiency and accuracy.

		\subsection{Action Recognition from Point Cloud Videos}
		An early work on human action recognition by Rahmani et al. use features derived from histograms of oriented 4 dimensional surface normals (HON4D) \cite{HON4d}, accumulating projections of unit surface normals onto vertices of a polychoron. Actions are classified with HON4D features using a support vector machine. You et al. present a real-time action recognition framework using a multi-view RGB-D setup, Action4D \cite{Action4d}. A 3D point cloud is built from calibrated RGB-D images, segmented into occupancy voxels, and processed with a 3D CNN for person detection. Candidate person volumes are identified using a 3D CNN input with bird's-eye view images, refined with Gaussian filtering and non-maximal suppression (NMS), and tracked using features like Euclidean distance, occupancy differences, and classification probabilities. Features are extracted from person voxels with a 3D CNN, where temporal dynamics are modeled using a long-short term memory (LSTM) module.\\ \indent Wang et al. author the 3DV (3D dynamic voxel) framework \cite{3DV}, using occupancy grids over time. Two streams are used, a motion stream consisting of voxels whose features consist of grid locations and a motion vector constructed using temporal rank pooling \cite{TemporalRankPooling}, input to a PointNet++ model for feature extraction. A second stream uses PointNet++ enacting on raw occupancy grids at strided frame times, where both streams are concatenated to a final linear classifier for action recognition.\\ \indent In PSTNet (Point Spatio-Temporal Network) \cite{PSTNET}, Fan et al. propose a novel point spatio-temporal convolution operator. The method begins by selecting a set of anchor frames based on specified stride and padding values. Within these anchor frames, anchor points are determined using iterative farthest point sampling. For each anchor point, a factorized spatio-temporal convolution is applied. This involves identifying a radius-based neighborhood of points for each frame within a temporal window, where the anchor points are propagated across all frames in the window. Dynamic kernel weights are then computed based on the displacements of points within these neighborhoods. The proposed PST convolution layer is incorporated into architectures designed for 4D semantic segmentation and action recognition. \\ \indent Ben-Shabat et al. \cite{3DInAction} author the 3D-In-Action framework, using t-patches to capture spatio-temporal dynamics in point cloud sequences. A t-patch represents the K-nn around a point in a given frame, which is linked across frames by finding the nearest neighbor in the next frame to the previous t-patch center, linking region groupings over time. For a sequence of point clouds, t-patches are embedded via MLP layers applied to spatial features, followed by pooling across points. These embeddings are processed with convolutional layers that operate across temporal and feature dimensions, forming t-patch modules, yielding per-frame feature vectors. These per-frame vectors are pooled across frames with temporal smoothing to make frame-level predictions. The architecture features a hierarchical structure of three t-patch modules, where iterative farthest point sampling reduces the number of points by half at each level.
		\\ \indent Distinct from the approach proposed in this paper, the preceding models do not provide human segmentation in image space or 3D space, leaving a large number of irrelevant background points in the data, and moreover do not segment body parts, or use surface normals. Further, they all consider that depth is measured from a sensor, but never estimated with monocular depth estimation. The models detailed in the previous literature review do not explore hybrid sparse convolution and point-based approaches, hence the motivation to explore these alternatives.
	\section{Methodology}
	\subsection{Human Point Clouds From Aligned Depth/IR Sequences} 
	In order to obtain human point clouds over time, first consider the scenario of aligned pairs of equal-resolution images $(I_t, D_t)_{t=1}^T$ for a sequence of $T$ frames, where $D_t$ are depth images. In the case of RGB-D images, $I_t$ will be an RGB image. With respect to experiments conducted in Section IV, since the NTU RGB-D dataset does not provide aligned RGB/depth images, treat $I_t$ as the pixel aligned infrared (IR) images. It is assumed that camera intrinsics are available. An overview of the proposed pipeline is shown in Figure 1.  
	\begin{figure}
		\centering 
		\begin{adjustbox}{max width=\linewidth}
			\begin{tikzpicture}[
				node distance=2cm and 3cm, 
				every node/.style={draw, rectangle, minimum width=5cm, minimum height=4cm, align=center, fill=blue!5},
				every join/.style={->, thick},
				]
				\node (pairs) {
					\begin{tabular}{c}
						Aligned Depth/IR Pairs\\
						\includegraphics[width=6cm,height=5cm]{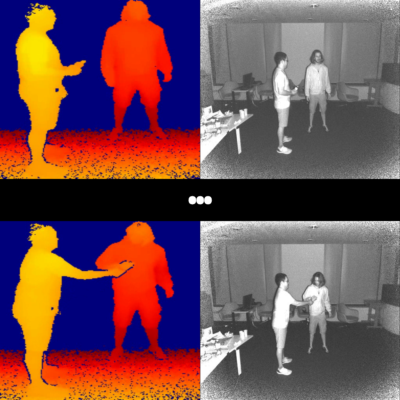} 
					\end{tabular}
				};
				\node[right=of pairs] (segmentations) {
					\begin{tabular}{c}
						Instance/Part Segmentations\\
						\includegraphics[width=6cm,height=5cm]{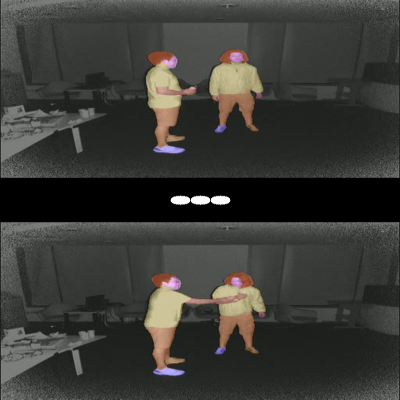} 
					\end{tabular}
				};
				\node[right=of segmentations] (denoising) {
					\begin{tabular}{c}
						Instance Mask Denoising\\
						\includegraphics[width=6cm,height=5cm]{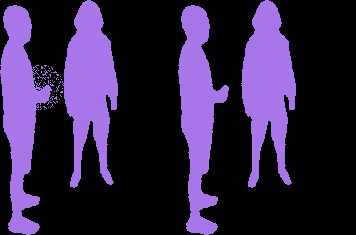} 
					\end{tabular}
				};
				\node[below=of pairs] (Person Tracking) {
					\begin{tabular}{c}
						Person Tracking\\
						\includegraphics[width=6cm,height=5cm]{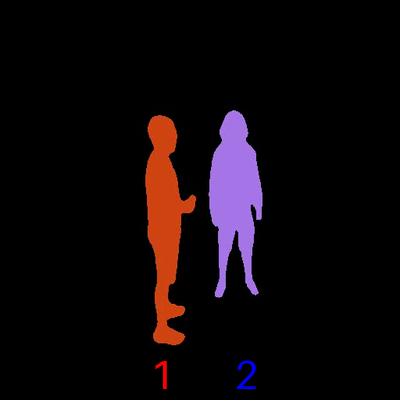} 
					\end{tabular}
				};
				\node[right=of Person Tracking] (projection) {
					\begin{tabular}{c}
						Projection to 3D\\
						\includegraphics[width=6cm,height=5cm]{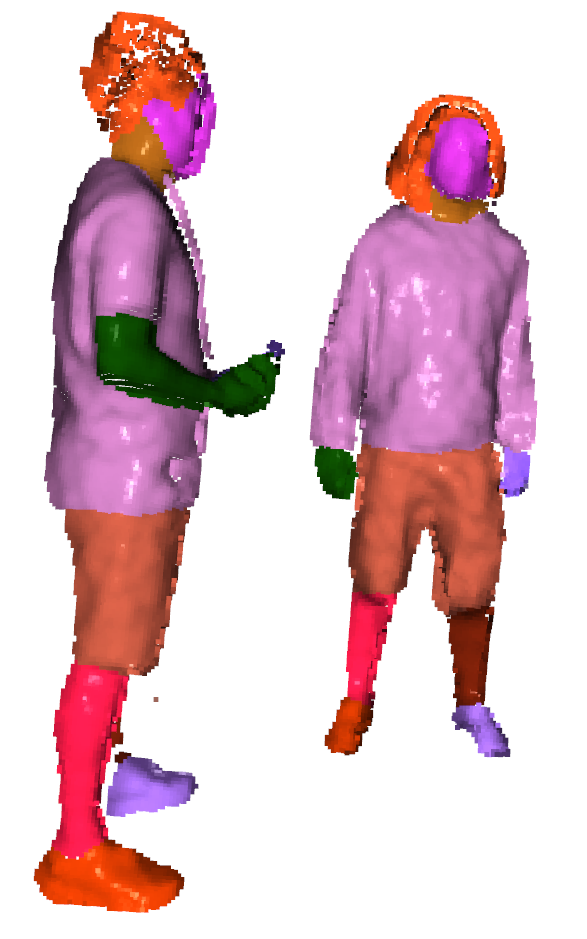} 
					\end{tabular}
				};
				\node[right=of projection] (pointcloud) {
					\begin{tabular}{c}
						Point Cloud Denoising\\
						\includegraphics[width=6cm,height=5cm]{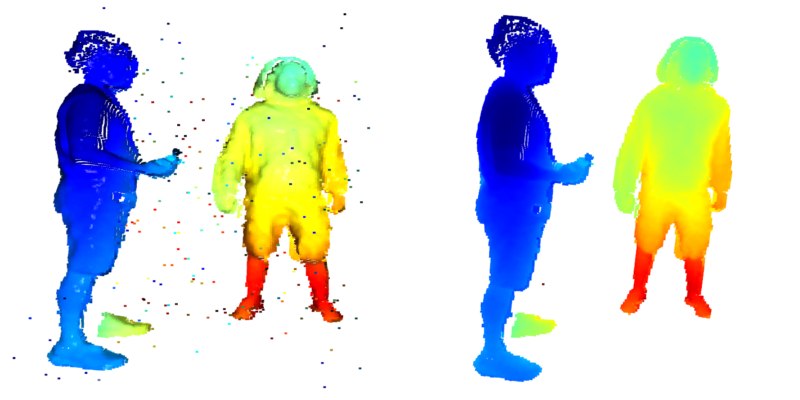} 
					\end{tabular}
				};
				
				\draw[->, line width=0.5mm, opacity=0.5] (pairs) -- (segmentations);
				\draw[->, line width=0.5mm, opacity=0.5] (segmentations) -- (denoising);
				\draw[->, line width=0.5mm, opacity=0.5] 
				(denoising.south west) -- 
				(Person Tracking.north east);
				\draw[->, line width=0.5mm, opacity=0.5] (Person Tracking) -- (projection);
				\draw[->, line width=0.5mm, opacity=0.5] (projection) -- (pointcloud);
			\end{tikzpicture}
		\end{adjustbox}
		\caption*{\textit{Figure 1: An overview of the proposed pipeline for obtaining human point clouds over time from aligned depth/IR pairs. Instance and body part segmentation is performed at each frame, followed by mask denoising in image space. Tracking is then applied to bounding boxes fit to the instances, followed by projection to 3D using the depth images, and denoising in 3D space. Note that body part labels are shown in the middle column.}} 
		\label{fig:pipeline}
	\end{figure}
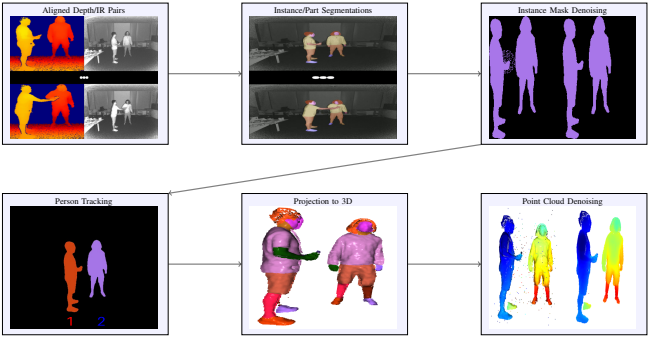
	Initially, simultaneous instance and body part segmentation, often referred to as human parsing, is performed on each individual image $I_t$ in a sequence. For this, the high performing model M2FP (Mask2FormerParsing) \cite{M2FP} was employed, based off the Mask2Former architecture \cite{M2F}, which computes instance, body part, and background masks from learnable mask queries input to a Transformer decoder. Instance masks are then denoised with simple heuristics, including removing very small connected components, and filling in the convex hull of instance masks to remove small interior holes caused by parsing inaccuracy. The result for frame $t$ is a set of $N$ instance masks $M_{t, n} \in \{0, 1\}^{H \times W}$ and per-image body part masks $P_t \in \{0, C+1 \}^{H \times W}$ for $C$ non-background classes, with $0$ given to the background label. \\
	\indent Using these per-frame instances with corresponding depth images, each person instance is projected to 3D considering the camera focal length $f$ and principal point $(c_x, c_y)$, and using the point cloud projection equations from a point $(x, y)$ in image space with depth value $z$ as in
	\begin{equation} X =   \frac{(x - c_x) * z}{f} , 
	Y =   \frac{(y - c_y) * z}{f}, 
	Z = z 
	\end{equation} 
	\indent Small clusters in 3D are removed with the DBSCAN algorithm \cite{DBSCAN}, and points more than 1.5 meters away from the body centroid, i.e. the projected instance mask centroid, are pruned. Invalid depth pixels are ignored. Using the tight-fitting bounding boxes fit to the instance segmentation masks per person, tracking is computed using the ByteTrack algorithm \cite{ByteTrack}. Sampling of points from each person is done for each frame using iterative farthest point sampling (IFPS) \cite{PointNet++} to a fixed size, typically either 1024 or 2048 points. Surface normals are computed per point, estimated with local plane-fitting. 
	
	\subsection{Human Point Clouds from RGB using Monocular Depth Estimation}
	It is often the case that the input to an action recognition algorithm is an RGB video from a single view, with no depth sensing technology present. Hence, capturing 3D geometry in this scenario is challenging. However, recent development of algorithms for monocular depth estimation has seen remarkable progress, where for a given color image, a disparity map is estimated. Modern approaches often train on mixtures of real-world datasets obtained with depth sensors, and photorealistic synthetic images rendered in computer graphics engines, focusing on fine-grained structural details as well as ordinal relations between points. In this work, the Depth-Anything v2 (DAv2) approach was chosen for its fine-grained accuracy, crucial for action recognition \cite{DepthAnything}. An overview of the proposed pipeline in this setting is shown in Figure 2. 
	\begin{figure}[H]
		\centering 
		\begin{adjustbox}{max width=\linewidth}
			\begin{tikzpicture}[
				node distance=2cm and 3cm, 
				every node/.style={draw, rectangle, minimum width=5cm, minimum height=4cm, align=center, fill=blue!5},
				every join/.style={->, thick},
				]
				\node (Pairs) {
					\begin{tabular}{c}
						RGB Frames and Estimated Disparity Maps\\
						\includegraphics[width=6cm,height=6cm]{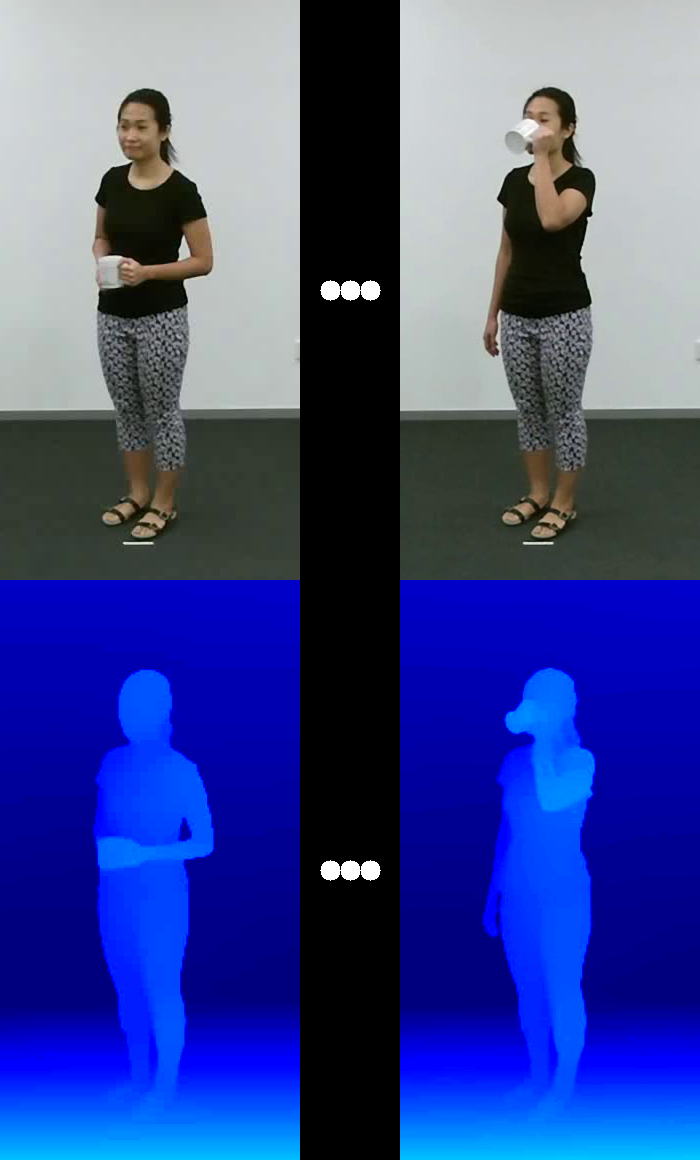} 
					\end{tabular}
				};
				\node[right=of pairs] (Segmentations) {
					\begin{tabular}{c}
						Instance/Part Segmentations\\
						\includegraphics[width=6cm,height=6cm]{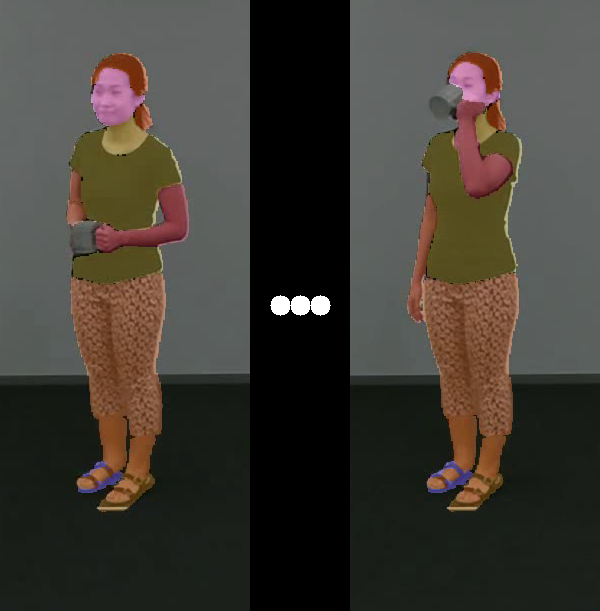} 
					\end{tabular}
				};
				\node[right=of segmentations](IMDenoising) {
					\begin{tabular}{c}
						Instance Mask Denoising\\
						\includegraphics[width=6cm,height=6cm]{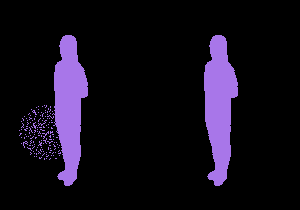} 
					\end{tabular}
				};
				\node[below=of pairs] (PersonTracking) {
					\begin{tabular}{c}
						Person Tracking\\
						\includegraphics[width=6cm,height=6cm]{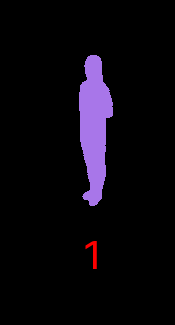} 
					\end{tabular}
				};
				\node[right=of PersonTracking] (Projection) {
					\begin{tabular}{c}
						Projection to 3D\\
						\includegraphics[width=6cm,height=6cm]{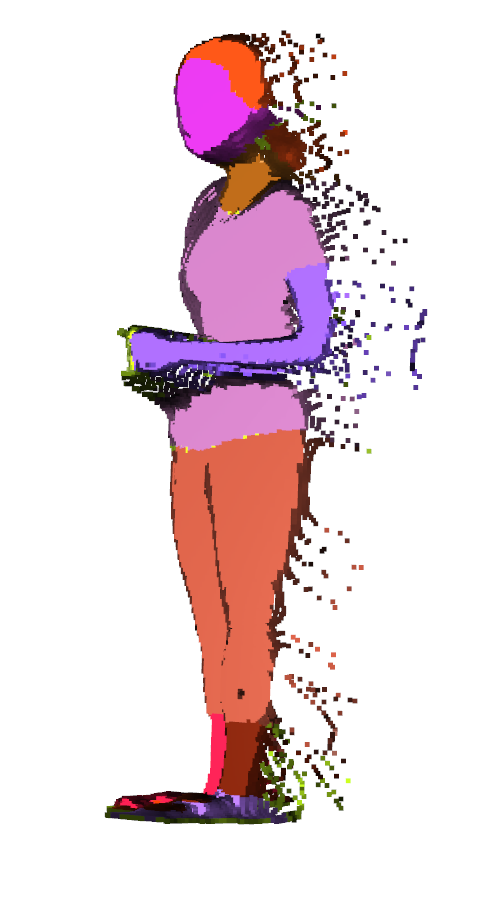}
					\end{tabular}
				};
				
				\node[right=of Projection] (PCDenoising) {
					\begin{tabular}{c}
						Point Cloud Denoising\\
						\includegraphics[width=6cm,height=6cm]{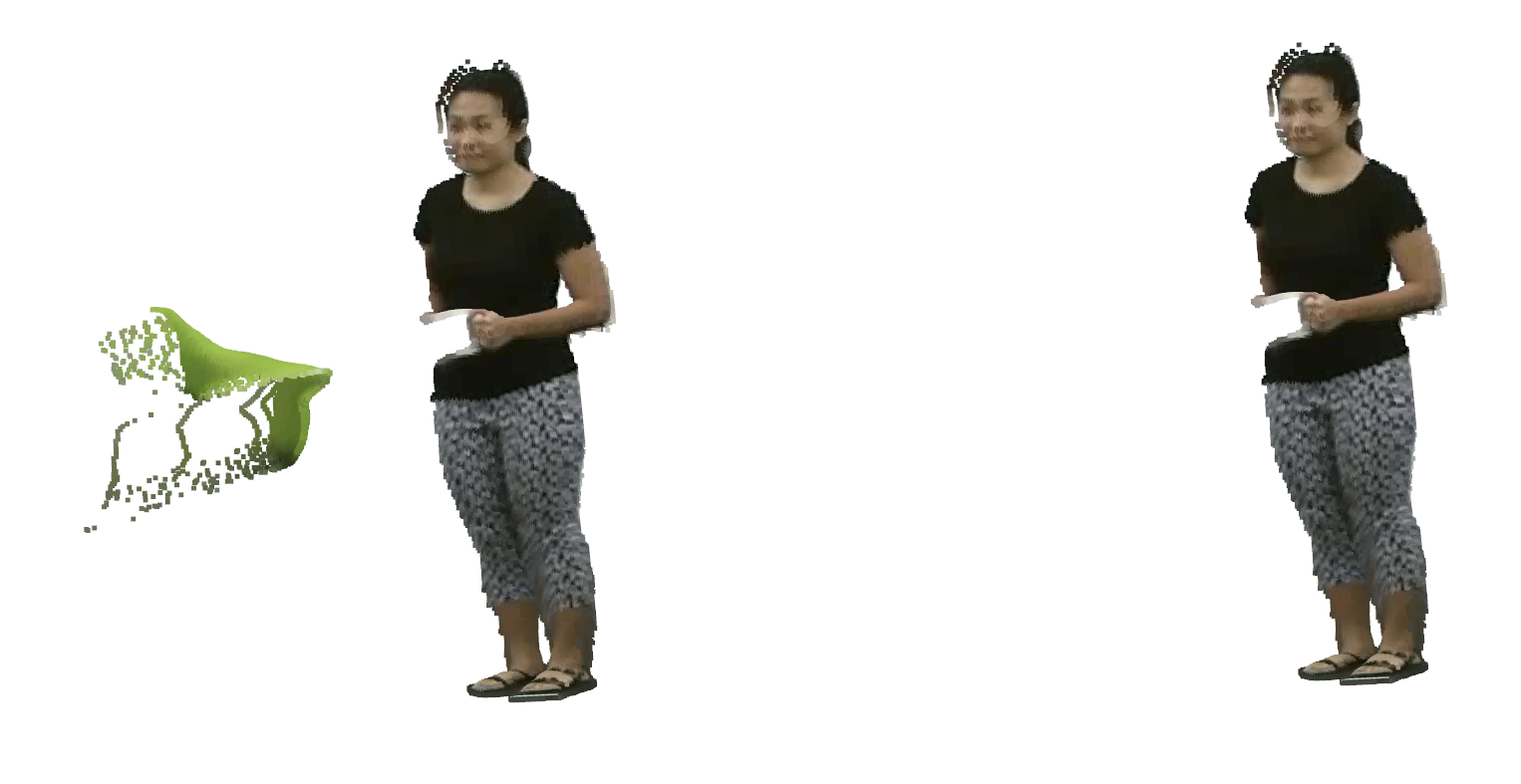}
					\end{tabular}
				};

				\draw[->, line width=0.5mm, opacity=0.5] (Pairs) -- (Segmentations);
				\draw[->, line width=0.5mm, opacity=0.5] (Segmentations) -- (IMDenoising);
				\draw[->, line width=0.5mm, opacity=0.5] 
				(IMDenoising.south west) -- 
				(PersonTracking.north east);
				\draw[->, line width=0.5mm, opacity=0.5] (PersonTracking) -- (Projection);
				\draw[->, line width=0.5mm, opacity=0.5] (Projection) -- (PCDenoising);
			\end{tikzpicture}
		\end{adjustbox}
		\caption*{\textit{Figure 2: The pipeline for obtaining denoised human point clouds over time from RGB videos using monocular depth estimation, with body part labels displayed in the middle column.}} 
		\label{fig:pipeline}
	\end{figure}
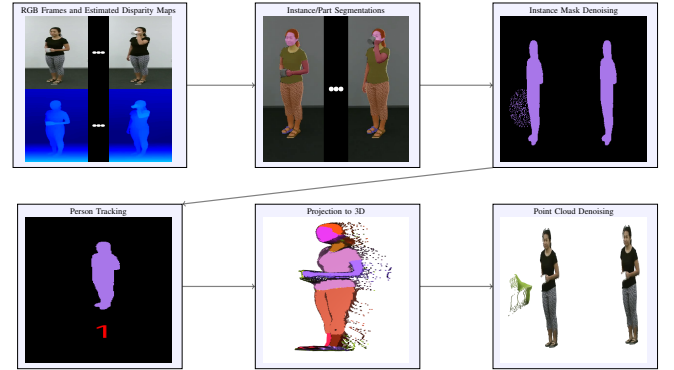 
	\indent Given a sequence of RGB frames $I_t$, metric disparity maps are estimated with DAv2, yielding $D'_t$, where using a scaling factor $S$, estimated depth maps are obtained as \begin{equation} D''_t (x,y) = \frac{S}{\epsilon + D'_t(x, y)} \end{equation} where $\epsilon $ is a small constant used to prevent division by zero. Denoising of instance masks, in addition to denoising in 3D is similar to section III-A, using DBSCAN and a percentile-based denoising strategy rather than a metric distance. In particular, points that are outside percentile range of $5-90$ of distance from the projected instance mask centroid are pruned. Since RGB images often have much higher resolution that depth images, sampling was applied to windows of points sorted by serialization with space-filling curves \cite{PointTransformerv3}, reducing the complexity for $K$ sample points, $W$ windows, and $N$ overall points from $O(NK)$ to $O(NK/W)$. Surface normals are also estimated in this setting. 
	\subsection{Action Recognition From Human Point Clouds over Time}
		In this section, the proposed action recognition model is detailed. It is now assumed that, given the denoised output from sections III, the input to the model are batches of tensors $S$ of shape $(B, T, N, C)$, for $T$ input frames, $N$ points, and $C$ channels, and batch size $B$. In the setting  of point cloud sequences obtained with depth sensors, channels consist of $3D$ coordinates, and possible auxiliary features, namely surface normals, infrared intensity, and part label features. Part labels are converted into learnable embeddings of dimension 3 using an embedding table. In the setting of point clouds estimated with monocular depth estimation, identical features are used, replacing infrared with RGB values. An overview is shown in Figure 3.
	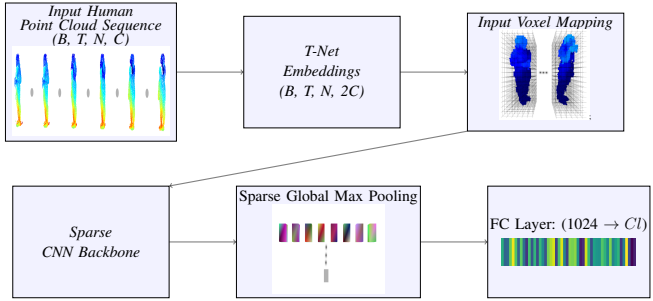
\begin{figure}[H]
	\usetikzlibrary{positioning}
		\begin{adjustbox}{max width=\linewidth}
		\begin{tikzpicture}[
			node distance=2cm and 3cm, 
			every node/.style={draw, rectangle, minimum width=7cm, minimum height=5cm, align=center, fill=blue!5},
			every join/.style={->, thick},
			]
			
			\node (Node1) {\begin{tabular}{c}
					\huge \textit{Input Human} \\
					\huge \textit{Point Cloud Sequence}\\
					\huge \textit{(B, T, N, C)} \\
					\includegraphics[width=7cm,height=4cm]{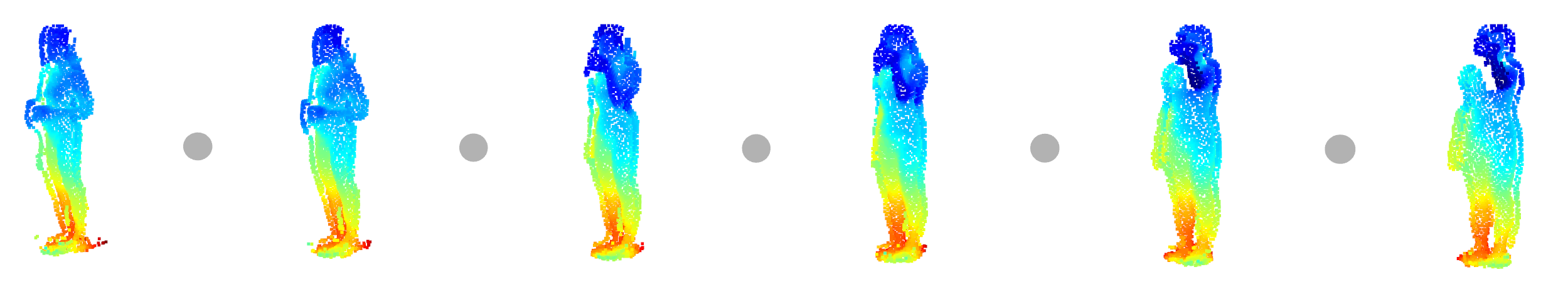}
			\end{tabular}};
			\node[right=of Node1] (Node2) {\huge \textit{T-Net}\\\\ 
				\huge \textit{Embeddings} \\\\
				\huge \textit{(B, T, N, 2C)} };
			\node[right=of Node2] (Node3) {
							\huge \textit{Input Voxel Mapping}\\
				{\includegraphics[width=4cm, height=4cm]{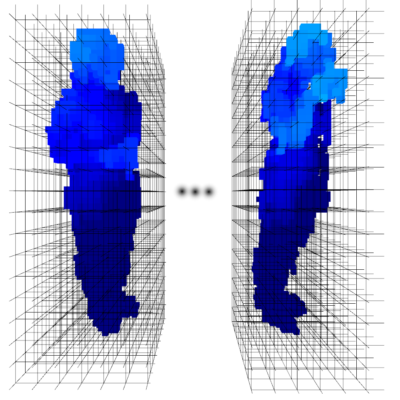}};\\
			};
			
			\node[below=of Node1] (Node4) {\huge \textit{Sparse} \\\\ \huge \textit{CNN Backbone}};
			\node [right= of Node4] (Node5) {\huge Sparse Global Max Pooling\\
				\includegraphics[width=5cm, height=4cm]{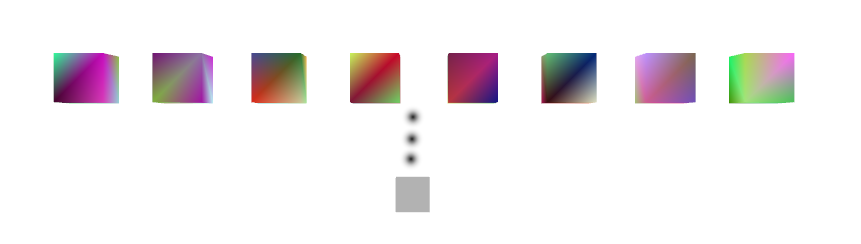}};
			\node [right= of Node5] (Node6) 
			{\huge FC Layer: (1024 $\rightarrow Cl$)\\\\
				\includegraphics[width=6cm, height=1.2cm]{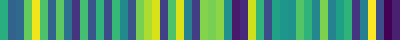}
			};
			
			\draw[->, line width=0.5mm, opacity=0.5] (Node1) -- (Node2);
			\draw[->, line width=0.5mm, opacity=0.5] (Node2) -- (Node3);
			\draw[->, line width=0.5mm, opacity=0.5] (Node3.south west) -- (Node4.north east);
			\draw[->, line width=0.5mm, opacity=0.5] (Node4) -- (Node5);
			\draw[->, line width=0.5mm, opacity=0.5] (Node5) -- (Node6);

		\end{tikzpicture}
		\end{adjustbox}
		\caption*{\textit{Figure 3: The proposed model for action recognition from human point clouds over time, where $T$ is the number of frames, $N$ the numbers of points, $C$ the number of input channels, $Cl$ the number of action class labels, and $B$ is the batch size.}} 
	\end{figure}
	 The overall architecture consists of a frame-wise T-Net (Transformation Net) embedding layer, followed by a sparse convolutional neural network, global sparse max pooling in voxel-space, and a fully connected layer for classification, to be trained with a standard cross-entropy loss.\\
	\subsubsection{T-Net Embeddings}
	In order to obtain per-frame, per-point embeddings, in this work the T-Net model is adopted from PointNet \cite{PointNet}. The T-Net module predicts a transformation matrix \( T \in \mathbb{R}^{C \times C} \) for global aggregation of features from point clouds. Given an input \( \mathbf{x} \in \mathbb{R}^{B' \times N \times C} \), with batch size $B' = B*T$ consisting of all individual frames in a sequence flattened over all sequences, the input is first transposed to \( \mathbf{x}^\top \in \mathbb{R}^{B' \times C \times N} \). Then, a series of multi-layer perceptron (MLP) operations with batch normalization (BN) and ReLU activations are applied, using weight matrices and bias values $W_i, b_i$ for $1 \leq i \leq 6$:
	\[
	\mathbf{h}_1 = \text{ReLU}(\text{BN}(\text{Conv1D}(\mathbf{x}^\top; W_1, b_1))) \quad \text{\(( B' \times 64 \times N \))}
	\]
	\[
	\mathbf{h}_2 = \text{ReLU}(\text{BN}(\text{Conv1D}(\mathbf{h}_1; W_2, b_2))) \quad \text{\( (B' \times 128 \times N \))}
	\]
	\[
	\mathbf{h}_3 = \text{ReLU}(\text{BN}(\text{Conv1D}(\mathbf{h}_2; W_3, b_3))) \quad \text{ \( (B' \times 1024 \times N \))}
	\]
	A global feature vector is obtained using max pooling:
	\[
	\mathbf{g} = \max_{n=1,\dots,N} \mathbf{h}_3[:, :, n] \quad \text{(\( B' \times 1024 \))}
	\]
	\\
	Three fully connected layers are used to predict a flattened transformation matrix: 
	\[
	\mathbf{z}_1 = \text{ReLU}(\text{BN}(\text{FC}(\mathbf{g}; W_4, b_4))) \quad \text{(\( B' \times 512 \))}
	\]
	\[
	\mathbf{z}_2 = \text{ReLU}(\text{BN}(\text{FC}(\mathbf{z}_1; W_5, b_5))) \quad \text{(\( B' \times 256 \))}
	\]
	\[
	T_\text{flat} = \text{FC}(\mathbf{z}_2; W_6, b_6) \quad \text{(\( B' \times (C \cdot C) \))}
	\]
	The flattened matrix \( T_\text{flat} \) is reshaped into \( T \in \mathbb{R}^{B' \times C \times C} \)
	The transformation matrix is initialized as:
	\begin{equation}
	T_\text{final} = T + \mathbf{I}_C,
	\end{equation}
	where \( \mathbf{I}_C \) is the \( C \times C \) identity matrix.
	
	The output \( T_\text{final} \in \mathbb{R}^{B' \times C \times C} \) is used to transform the input point cloud using batch matrix multiplication, after which the original features are concatenated to retain both local and global information, i.e. 
	\begin{equation} y = x^T||(T\circ x^T)
		\end{equation}
		where $\circ$ denotes batch matrix multiplication and $||$ denotes the concatenation operator, applied along the channel axis. The output is subsequently reshaped to $(B, T, N, 2C)$. T-Net embeddings allow the model, at each frame, to obtain global point cloud information, which is difficult to obtain with sparse convolution without a large receptive field, requiring many layers or large kernel sizes. In the next model stage, voxel mapping is applied to the first three channels of $y$ corresponding to the original 3D coordinates.\\
	\subsubsection{Sparse CNN Backbone}
	Before presenting the proposed sparse CNN backbone, the sub-manifold sparse convolution \cite{SubManifoldConvolution}, in 3D, is reviewed for completeness. The sub-manifold variant was chosen due to its computational efficiency, reducing the dilution effect of regular sparse convolution, restricting to sparse max pooling operations. For a grid size $(g_x, g_y, g_z)$,  each point $p=(p_x, p_y, p_z)$ in each frame is mapped to a voxel location $(x', y', z') \in \mathbb{Z}_{+}^3$ by coordinate division by the grid size, rounding down. \\
	\indent For a set of voxel-mapped points $P = \{p_1, \ldots p_N \}$, and a multi-channel kernel $K$, for each point $p$ mapped to a given voxel grid location $(x', y', z')$, sparse sub-manifold convolution enacts on point $p$ with kernel $K$ for output channel $c'$ for $C$ input channels as
	\begin{equation}
		p_c' = \sum_{c=1}^{C} \sum_{k \in O(K)} K(k)_{c', c} * A((x', y', z') + k, c) 
	\end{equation}
	where $O(K)$ are the offset location of kernel $K$ with non-empty voxels relative to center location $(x', y', z')$, and $A$ is an aggregation operation for all points falling into a given voxel for channel $c$, typically the mean or max value. For all sparse convolutions in the following, the assume the sub-manifold formulation is assumed, with no dilation, and zero-padding voxels to preserve the input resolution.\\
	\indent The proposed backbone is detailed in Table 1, with layer descriptions to follow, where the corresponding architecture will be referred to as SP-HP-ConvoT (Sparse Convolution for Human Point Clouds over Time). Note that in Table 1, and Figure 4, output shapes are shown in dense voxelized form, but the sparse convolution data structure stores features per point, and their respective spatio-temporal grid and batch indices.
	\begin{table}[H]
		\begin{adjustbox}{max width=\linewidth}
			\renewcommand{\arraystretch}{2.2}
			\huge
			\begin{tabular}{|c|c|c|}
				\hline
				Layer Name & Output Tensor Shape & Layer Parameters/Description\\
				\hline
				Input & (B, T, N, 2C)  & Input Human Point Clouds over Time\\
				\hline
				Voxel Mapping & (B, T, $G_x, G_y, G_z$, 2C)  & Point to Grid Index Assignment \\
				\hline
				4D SubManifold Convolution Layer 1 & (B, T, $G_x, G_y, G_z$, 64) & Kernel: 5, Stride: 1, Padding: 3\\
				\hline
				4D Sparse Max Pooling & (B, T/2, $G_x/2, G_y/2, G_z/2$, 64) & Kernel: 3, Stride: 2, Padding: 1\\
				\hline
				4D SubManifold Convolution Layer 2 & (B, T/2, $G_x/2, G_y/2, G_z/2$, 128) & Kernel: (1, 7, 7, 7), Padding: (0, 3, 3, 3)\\
				\hline
				4D SubManifold MS-TCN & (B, T/2, $G_x/2, G_y/2, G_z/2$, 128) & Temporal Kernel Sizes: (3, 5, 7, 9)\\
				\hline
				4D Sparse Max Pooling & (B, T/4, $G_x/4, G_y/4, G_z/4$, 128) & Kernel: 3, Stride: 2, Padding: 1\\
				\hline 
				4D SubManifold Bottleneck Layer 1 & (B, T/4, $G_x/4, G_y/4, G_z/4$, 256) & BottleNeck dimension: 64\\
				\hline
				4D SubManifold Bottleneck Layer 2 & (B, T/4, $G_x/4, G_y/4, G_z/4$, 1024) & BottleNeck dimension: 128\\
				 \Xhline{8\arrayrulewidth}
				Sparse Global Max Pooling & (B, 1024) & Sparse Pooling along $t$, $x$, $y$ and $z$ \\
				\hline
				FC Classification Head & (B, Cl) & Single Fully Connected Layer\\
				\hline
			\end{tabular}
		\end{adjustbox}
		\caption*{\textit{Table 1: The SP-HP-ConvoT backbone, with sub-manifold convolutions, followed by sparse max pooling and a fully connected classification head. When a kernel, stride, or padding value is a single integer $n$, it should be interpreted as a shape broadcast to 4 dimensions, i.e. ($n, n, n, n)$. $B$ denotes the batch index, $T$ the number of frames, and $N$ denotes the number of points in the point cloud. $C$ is the number of input channels, equal to $3$ for purely geometric point clouds, and $Cl$ is the number of class labels.}}
	\end{table}
	
	Each point within each batch in the input is mapped to a 4-dimensional voxel grid index, $(T, G_x, G_y, G_z)$ using the first three channels of $y$ and the time axis dimension, described in Section III-C1. 
	The first layer employs a large kernel with an extent of $5$ across all dimensions, including time, to effectively capture and embed local relationships. Unlike other approaches, this network utilizes early spatio-temporal sub-manifold convolution, despite its relatively high computational cost, due to the significant performance gains it provides in the early stages. This design is similar to Minkowski Networks but features a slightly larger kernel size. This contrasts with approaches like Action4D \cite{Action4d} that do not directly model spatio-temporal relations with raw geometry, but extract per-frame embeddings. The first layer is followed by sparse max pooling \cite{MinkowskiEngine} to reduce the voxelized spatial extent, balancing the computation burden of dilution of features from sparse regions into their neighbors. Note that sparse max pooling performs full voxel aggregation, resulting in unique per-voxel features. \\
	\indent Inspired by factorized spatio-temporal convolutions in the 3D CNN architecture ResNet (2+1)-D \cite{ResNet2p1D}, and the skeletal-action recognition model ST-GCN++ \cite{St-GCN++}, a factorized spatio-temporal block is used, computing first a purely spatial convolution in SubManifold Convolution Layer 2. This is followed by a novel 4D sub-manifold multi-scale temporal convolutional network (MS-TCN), computed now in 4 dimensions. A depiction of the proposed 4D sub-manifold MS-TCN is shown in Figure 4.	The MS-TCN layer simply enacts a set of varying kernel sizes on the temporal axis per each non-empty voxel location, projected to a lower dimensional feature space, concatenating the results along the channel axis.
		
	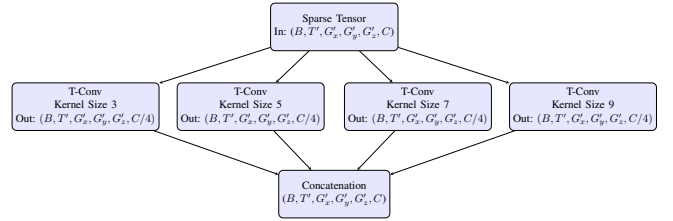
\begin{figure}[t]
		\centering
		\begin{adjustbox}{max width=\linewidth} 
			\begin{tikzpicture}[node distance=1.5cm and 1.5cm, >=stealth]
				
				\tikzstyle{block} = [rectangle, draw, fill=blue!10, text centered, rounded corners, minimum height=1.5cm, minimum width=2.5cm]
				
				\node[block, align=center] (main) {Sparse Tensor \\In: ($B, T', G'_x, G'_y, G'_z, C$)};
				
				\node[block, align=center, below left=1cm and 3.5cm of main] (block1) {T-Conv \\
				Kernel Size 3\\
				Out: $(B, T', G'_x, G'_y, G'_z, C/4)$};
				\node[block, align=center, below left=1cm and -1.7cm of main] (block2) {T-Conv \\
					Kernel Size 5\\
					Out: $(B, T', G'_x, G'_y, G'_z, C/4)$};
				\node[block, align=center, below right=1cm and -1.7cm of main] (block3) {T-Conv \\
					Kernel Size 7\\
					Out: $(B, T', G'_x, G'_y, G'_z, C/4)$};
				\node[block, align=center, below right=1cm and 3.5cm of main] (block4) {T-Conv \\
					Kernel Size 9\\
					Out: $(B, T', G'_x, G'_y, G'_z, C/4)$};
				
				\node[block, align=center, below=2cm of $(block2)!0.5!(block3)$] (final) {Concatenation\\
					($B, T', G'_x, G'_y, G'_z, C$)
				};
				
				\draw[->] (main) -- (block1);
				\draw[->] (main) -- (block2);
				\draw[->] (main) -- (block3);
				\draw[->] (main) -- (block4);
				
				\draw[->] (block1) -- (final);
				\draw[->] (block2) -- (final);
				\draw[->] (block3) -- (final);
				\draw[->] (block4) -- (final);
				
			\end{tikzpicture}
		\end{adjustbox}
		\caption*{\textit{Figure 4: The sub-manifold multi-scale temporal convolutional layer (MS-TCN), consisting of purely temporal convolutions acting on 4D voxelized space at various kernel sizes, with input voxel resolution $(T', G'_x, G'_y, G'_z)$}.}
	\end{figure}
	The final two convolutional layers are inspired by the standard residual bottleneck layers of the ResNet architecture \cite{ResNet}, but applied in 4D sub-manifold fashion and with a kernel of extent 3 in all axes. The channel matching layer consists of a point-wise convolution to match the output channels for residual addition, as shown in Figure 5.
	\begin{figure}[H]
		\centering
		\begin{adjustbox}{max width=\linewidth}
	\begin{tikzpicture}[node distance=5.0cm, align=left, font=\small]
		\tikzstyle{block} = [rectangle, draw, fill=blue!5, 
		text width=4.5cm, text centered, rounded corners, minimum height=2em]
		\tikzstyle{arrow} = [thin,->,>=stealth]
		
		\node [block, anchor=west] (submanifold1) {Sub-Manifold Conv Layer: \\ Kernel: 1, Padding : 0 \\ Stride 1 \\ In Channels : \textit{In}, Out Channels: \textit{In/2}, \\ BN, ReLU};
		\node [block, right of=submanifold1] (submanifoldbottleneck1) {Bottleneck Sub-Manifold Conv Layer\\ Kernel: 3, Padding : 1 \\ Stride: 1 \\ In Channels : \textit{In/2}, Out Channels: \textit{In/2}\\ BN, ReLU};
		\node [block, right of=submanifoldbottleneck1] (submanifold2) {Sub-Manifold Conv Layer\\ Kernel: 1, Padding : 1 \\ Stride 1 \\ In Channels : \textit{In/2}, Out Channels: \textit{Out}\\ BN, ReLU};
		\node [block, right of=submanifold2, xshift=0.9cm] (Identity) {(Optional) Channel Matching Sub-Manifold Convolution \\ $\rightarrow$ Element-wise Addition};
		\draw [arrow] (submanifold1) -- (submanifoldbottleneck1);
		\draw [arrow] (submanifoldbottleneck1) -- (submanifold2);
		\draw [arrow] (submanifold2) -- (Identity);
		\draw [arrow] (submanifold1.north east) to[bend left=40] (Identity.north west);
		\draw [arrow] (Identity.east) -- ++(1,0); 
	\end{tikzpicture}
	} 
	\end{adjustbox}
	\caption*{\textit{Figure 5: The 4D Sub-Manifold Residual Bottleneck Layers. Note that kernel sizes, stride, and padding values are broadcast to 4 dimensions. }}
	\end{figure}
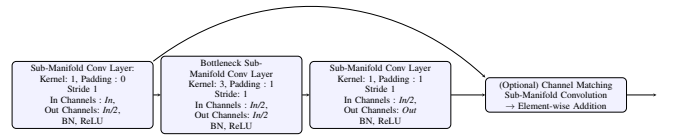 
	Sparse max pooling is subsequently applied across the channels of non-empty voxels, followed by a fully connected classification head. For multi-person action recognition, features are computed separately per each person, and averaged before the fully connected layer. 
	\section{Experiments}
	\subsection{Dataset}The NTU (Nanyang Technological University) RGB-D 120 dataset \cite{NTURGBD120} is a challenging multi-modal human action recognition dataset, popularly used for validation of skeletal action recognition models using 120 action classes. The dataset is obtained in indoor settings, filmed using a Kinect v2 time-of-flight depth camera. Using 106 different actors there are 114,480 video sequences in total. The sequence lengths vary from $15$ to $300$ frames. Videos were recorded from 32 collection setups, using varied locations and background views. In each setup, three different camera views are employed, using horizontal angles $\theta \in \{ -45, 0, 45 \}$, in degrees.\\
\indent Two splits are typically considered for model validation. In the \textit{cross-subject} (CSub) split, the training and testing data do not share the same actors, with $53$ training subjects, and $53$ test subjects, yielding $63,360$ and $ 51,120$ samples respectively. The second split is the \textit{cross-setup} (CSet) split, where half of the $32$ setups are used for training, and the other half for evaluation, with $54,720$ and $59,760$ samples each. As per the reported accuracy of modern skeletal action recognition models, the cross-subject split is the more challenging scenario \cite{St-GCN++}.
	\subsection{Implementation}
	Data pre-processing follows the procedure described in Section III A-B. Instance and body part segmentation were computed with the M2FP algorithm, using weights pre-trained on the Crowd Instance-level Human Parsing dataset (CIHP)\cite{CIHP}, which consists of 20 part labels: \textit{background, face, hair, torso skin, upper clothes, right arm, left arm, pants, coat, left shoe, right shoe, right leg, left leg, hat, dress, socks, sunglasses, skirt, scarf, and glove}. Infrared images are normalized using min-max normalization, preceded by clamping at the 5th and 90th percentiles, reproducing the intensity channel 3 times to form an RGB image. Point clouds are sampled to 2048 points per frame using the iterative farthest point sampling algorithm \cite{PointNet++} for point clouds obtained with a depth sensor, and the windowed approach described in Section III-B for point clouds obtained with monocular depth estimation, as the RGB resolution of the Kinect v2 is much larger than the depth image resolution. All point cloud processing was done offline, including surface normal computation.\\
	\indent During training, 32 frames per sequence were sampled using uniform random sampling, dividing the total number of frames into equally sized intervals and randomly sampling within those intervals, and duplicating frames for shorter sequences as needed. During inference single sequence testing was used, sampling frames uniformly. All experiments were conducted using the PyTorch library with a batch size of $16$, split between 2 Nvidia RTX 4090 GPUs. The learning rate was set initially to $10^{-3}$, using the AdamW optimizer, and weight decay regularization of $10^{-5}$.  The learning rate was decayed to a minimum value of $10^{-8}$ evenly across $60$ epochs. A precision of 16-bits was used for efficiency of computation and memory usage. Sparse convolutions are computed with the Spconv library \cite{SPConvLibrary}.\\
	\indent Data augmentation was used during training, consisting of random rotations about the y-axis sampled uniformly in the range $\theta \sim U[-\pi/4, \pi/4]$, as well as point jittering. Sequences were normalized using coordinate-wise min-max normalization per person, applied using minimum and maximum $(x,y,z)$ coordinates over the entire sequence. Sequences in the NTU RGB-D 120 dataset are either single or two person actions. For two-person sequences, the second person's point cloud features are treated as a separate batch index and concatenated at the backbone output before the network head. For single-person actions, a zero tensor of dimension $ d= 1024$ is concatenated at the same stage. Unlike ST-GCN \cite{ST-GCN}, which zero-pads missing second-person inputs, the proposed approach computes features only for present actors, improving efficiency.
	
	\subsection{Results}
	In the following, SP-HP-ConvoT refer to models using point cloud sequences from a depth sensor, and SP-HP-ConvoT-MDE for point cloud sequences obtained from monocular depth estimation (MDE). In Table 2, the effects of the different auxiliary channels are considered, namely infrared, surface normals, part label embeddings, and color. 
	
	\begin{table}[H]
		\begin{flushleft} 
			\begin{adjustbox}{max width=\linewidth}
				\begin{tabular}{|l|l|}
					\hline
					Model Name & Accuracy (\%)\\
					\hline 
					SP-HP-ConvoT & 85.8 \\
					\hline 
					SP-HP-ConvoT with IR  & 87.7 \\
					\hline 
					SP-HP-ConvoT with Normals & 86.8 \\
					\hline 
					SP-HP-ConvoT with Parts & 86.3 \\
					\hline 
					SP-HP-ConvoT with IR + Normals + Parts & \textbf{88.0}  \\ 
					\hline
					\hline
					SP-HP-ConvoT-MDE &  78.7 \\
					\hline 
					SP-HP-ConvoT-MDE with RGB & 80.1 \\
					\hline 
					SP-HP-ConvoT-MDE with Normals & 78.9 \\
					\hline   
					SP-HP-ConvoT-MDE with Parts & 82.0 \\
					\hline 
					SP-HP-ConvoT-MDE with RGB + Normals + Parts & 83.8 \\
					\hline
				\end{tabular}
			\end{adjustbox}
		\end{flushleft}
			\caption*{\textit{Table 2: Comparing the influence of auxiliary features vs. accuracy on the CSub split of the NTU RGB-D 120 dataset.}}
	\end{table}
	Each feature adds a small but significant boost in performance versus the raw geometry, where for point clouds obtained from the Kinect v2 (SP-HP-ConvoT model), the infrared and surface normals contribute to the largest boost in accuracy. Parts estimated from infrared images may be less accurate than those on RGB images given the domain shift, which may explain the modest gain in performance from adding body part segmentation embeddings vs. the raw geometry. The point clouds obtained from monocular depth estimation (SP-HP-ConvoT-MDE model) perform worse in general, but benefit greatly from body part segmentation embeddings (+3.3\%).  
	In Table 3, the effect of the T-NET embeddings is compared to a light SP-HP-ConvoT model without an embedding stage, in addition to a model without the MS-TCN layer, replaced with a single temporal convolution of kernel size $7$. The benefits show modest improvements in performance for each, and in conjunction with each other (+1\%).
	\begin{table}[H]
		\small 
		\begin{adjustbox}{max width=\linewidth}
		\begin{tabular}{|l|c|c|c|}
			
			\toprule
			Model & T-Net Embeddings & MS-TCN  & Accuracy (\%) \\
			\midrule
			SP-HP-ConvoT light & \xmark & \xmark & 84.8 \\
			+ T-Net Embeddings & \cmark & \xmark & 85.4 \\
			+ MS-TCN & \xmark & \cmark & 85.3 \\
			+ T-Net Embeddings + MS-TCN & \cmark & \cmark & \textbf{85.8} \\
			\bottomrule
		\end{tabular}
			\end{adjustbox}
				\caption*{\textit{Table 3: Comparing the influence of the proposed T-NET and MS-TCN modules vs. accuracy on the CSub split of the NTU RGB-D 120 dataset, using the SP-HP-ConvoT model.}}
	\end{table}
	Next, in Table 4, the proposed model is compared against three architectures from the skeletal action recognition literature (SAR). In this comparison comparisons were made from models that take either the raw keypoint representation, commonly referred to as \textit{joints}, or the \textit{bones} representation as input. The \textit{bones} input format consists of link vectors between keypoints, defined by an adjacency matrix, and is frequently used in \textit{joint-bone} ensembles, where predictions from separate models are combined with weighted contributions. The proposed approach is competitive, lightly outperforming existing reported methods on the CSet split.  
	
	\begin{table}[H]
		\begin{flushleft} 
			\begin{adjustbox}{max width=\linewidth}
				\begin{tabular}{|l|l|l|}
					\hline 
					Model Name & CSub Acc. (\%) & CSet Acc. (\%) \\
					\hline 
					ST-GCN++ \cite{St-GCN++} - Joints & 83.2 & 84.4 \\
					\hline
					ST-GCN++ \cite{St-GCN++} - Bones & 85.6 & 84.8 \\
					\hline
					InfoGCN \cite{InfoGCN} - Joints & 85.1 & 86.3 \\
					\hline 
					InfoGCN \cite{InfoGCN} - Bones & 87.3 & 88.5 \\
					\hline 
					DeGCN \cite{DGCN} - Joints & 87.6 & - \\
					\hline 
					DeGCN \cite{DGCN} - Bones & \textbf{88.5} & - \\
					\hline 
					SP-HP-ConvoT with IR + Normals + Parts & 88.0 & \textbf{88.7} \\
					\hline
					SP-HP-ConvoT-MDE with RGB + Parts + Normals & 83.8 & 84.8 \\
					\hline 					
				\end{tabular}
			\end{adjustbox}
		\end{flushleft}
		\caption*{\textit{Table 4: Comparing the performance of the proposed approach to representative models from the Skeletal Action Recognition literature. Note that DeGCN does not report the accuracy of models input with only joint or bones on the CSet split.}}
	\end{table}
	
	Inference speed is evaluated in Table 5 based on the number of points sampled from the point cloud and people, measured in sequences per second on a single Nvidia RTX 4090 GPU. The evaluation uses sequence lengths of 32 frames with a batch size of 1, considering only raw geometry (i.e., 3 input channels). For comparison, the SAR baseline ST-GCN++ \cite{St-GCN++} is included, using 133 keypoints from the COCO WholeBody dataset \cite{COCO_WB}, selected for its relatively dense keypoint representation, with either one or two people. While the proposed approach runs slower, it achieves 40 sequences per second with 2048 points when processing two people.
	
	\begin{table}
		\begin{flushleft} 
			\begin{adjustbox}{max width=0.9\linewidth}
				\begin{tabular}{|c|c|c|}
					\hline 
					Input Size & Number of People & Sequences per Second \\
					\hline
					SP-HP-ConvoT  & & \\ 
					\hline  
					512 points & 1 & 105.0 \\
					\hline 
					512 points & 2 & 85.6 \\
					\hline 
					1024 points & 1 & 85.6 \\
					\hline 
					1024 points & 2 & 60.3 \\
					\hline
					2048 points & 1 & 61.8 \\
					\hline 
					2048 points & 2 & 40.1	\\
					\hline 
					\hline 
					ST-GCN++ & & \\
					\hline 
					133 keypoints & 1 & 134.1 \\
					\hline 
					133 keypoints & 2 & 133.6 \\
					\hline 				
				\end{tabular}
			\end{adjustbox}
		\end{flushleft}
		\caption*{\textit{Table 5: Comparing the inference speed of SP-HP-ConvoT to a SAR baseline ST-GCN++ \cite{St-GCN++} in terms of sequences per second, versus the number of input points/keypoints and people.}}
	\end{table}
	\vspace{3px}
	In Table 6, the results of existing point cloud action recognition models to the proposed approach is compared, also considering an ensemble of the SP-HP-ConvoT and SP-HP-Convot-MDE models, using weights $\lambda_1, \lambda_2$ obtained through optimizing performance on a separate validation set during training, i.e. predictions of the form 
	{
	\begin{equation}
	y = \lambda_1 \text{SP-HP-ConvoT} (x) + \lambda_2 \text{SP-HP-ConvoT-MDE} (x)
	\end{equation}}
	assuming all auxiliary features for notational brevity.\\
	\indent 
	  On the CSub split, the SP-HP-ConvoT model with auxiliary features outperforms existing approaches by 1.0\%. The proposed ensemble achieves the best result (+2.3\%). However, both in the single model and ensemble setting on the CSet benchmark, performance lags behind. It may be that the accuracy loss may stem from the cumulative effect of person and body part segmentation inaccuracies. If the model has not encountered similar noise patterns or collection setups during training, which vary in the CSet setting, it may struggle to distinguish body points from the background as effectively.
	
	\begin{table}[H]
		\begin{flushleft} 
			\begin{adjustbox}{max width=\linewidth}
				\begin{tabular}{|l|l|l|}
					\hline 
					Model Name & CSub Acc (\%) & CSet Acc. (\%) \\
					\hline
					3DV \cite{3DV} & 82.4 & 93.5 \\
					\hline 
					PSTNet \cite{PSTNET} & 87.0 & \textbf{93.8} \\ 
					\hline 
					SP-HP-ConvoT with IR + Normals + Parts & 88.0 & 88.7 \\
					\hline 
					SP-HP-ConvoT-MDE with RGB + Normals + Parts & 83.8 & 84.8 \\
					\hline 
					\makecell[l]{SP-HP-ConvoT with IR + Normals + Parts \\ + SP-HP-ConvoT-MDE with RGB + Normals + Parts} & \textbf{89.3} & 91.2\\
					\hline	
				\end{tabular}
			\end{adjustbox}
		\end{flushleft}
		\caption*{\textit{Table 6: Comparing the performance of the proposed approaches to existing point cloud action recognition models from the literature.}}
	\end{table}

	\subsection{Discussion}
	In general, the model introduced in this work performs competitively with state-of-the art models both from the SAR and point cloud video deep learning approaches. In the depth sensor setting, segmentation accuracy may be less than optimal due to the domain shift of using a human parsing model trained on RGB images directly to normalized IR images. With respect to monocular depth estimation, the estimated body shapes may be too planar to distinguish certain motions, and suffer from reduced quality when subjects are viewed from the side. Further, for both the depth sensor and MDE setting, the worst performing classes were related to fine-grained hand movements, where in Table 7, 5 of the worst per-class $F_1$ scores are listed. Interestingly, Myung et al. observe a similar issue with regards to SAR models \cite{DGCN}, namely that failure cases tend to involve precise hand movements. 

\begin{table}
		\begin{flushleft}
		\begin{tabular}{|c|c|} 
			\hline
			Class Name & $F_1$ Score (\%) \\
			\hline
			make victory sign & 58.1 \\
			\hline
			staple book & 62.4 \\
			\hline
			cutting paper & 64.9 \\
			\hline
			make OK sign & 65.2\\
			\hline
			thumbs up & 72.4 \\
			\hline
		\end{tabular}
	\end{flushleft}
		\caption*{\textit{Table 7: The $F_1$ scores of the worst performing classes for the SP-HP-ConvoT + SP-HP-ConvoT-MDE ensemble.}}
\end{table}
	
	As a possible explanation for such confusion in between classes, the iterative farthest point sampling algorithm may over-sample parts of the torso and legs, and under-sample the arms and hands of subjects. Future work should remedy this by explicitly segmenting hands in addition to body parts, and either developing a separate hand model, or over-sampling hand points, to maximize performance. 

	\section{Conclusion}
	In conclusion, this work presents a novel approach to human action recognition from point cloud sequences. A pipeline for segmenting body parts and tracking individuals was employed, leveraging 3D projection with depth images and, alternatively, monocular depth estimation. A novel backbone using sparse convolutional neural networks was developed, one of the first of its kind in the literature for human action recognition from point clouds. Model performance was validated on a popular action recognition dataset, showing competitive results and a new benchmark high for temporal point cloud deep learning on the NTU RGB-D 120 dataset within the CSub split. \\
	\indent In future work, the aim will be to refine the segmentation algorithm with fine-tuning in the IR setting, as well as explore more modern approaches to human-based monocular depth estimation. Efforts to maintain efficiency of the entire pipeline, as well as model inference should be researched. Further, action recognition more centered on explicitly sampling of points on the hands and arms of individuals may yield improved accuracy. 

	\section*{Acknowledgment}
	This research was supported in part by MITACS
	Accelerate grant \#IT29551.
	
	\clearpage
	\bibliographystyle{IEEEtran} 

\end{document}